\documentclass[preprint]{elsarticle}

\usepackage{times}  
\usepackage{helvet}  
\usepackage{courier}  
\usepackage{url}  
\usepackage{graphicx}  

\usepackage{latexsym}
\usepackage{amsmath}
\usepackage{multirow}
\usepackage{enumerate}
\usepackage{amssymb}
\usepackage{amsfonts}

\usepackage{caption}
\usepackage{subfigure}
\usepackage[hidelinks, colorlinks, linkcolor=red, anchorcolor=purple,citecolor=blue]{hyperref}
\usepackage[linesnumbered,boxed]{algorithm2e}
\usepackage{xspace,mfirstuc,tabulary}

\journal{Arxiv}

\begin{document}
	
	\begin{frontmatter}
		
		
		\title{Synchronous Bidirectional Inference for Neural Sequence Generation}

		\author[casia,gucas]{Jiajun Zhang\corref{correspondingauthor}}
		\ead{jjzhang@nlpr.ia.ac.cn}
		
		\author[casia,gucas]{Long Zhou}
		\ead{long.zhou@nlpr.ia.ac.cn}
		
		\author[casia,gucas]{Yang Zhao}
		\ead{yang.zhao@nlpr.ia.ac.cn}
		
		\author[casia,gucas,bsit]{Chengqing Zong}
		\ead{cqzong@nlpr.ia.ac.cn}
		
		\cortext[correspondingauthor]{Corresponding Author}
		
		\address[casia]{National Laboratory of Pattern Recognition, CASIA, Beijing, China}
		\address[gucas]{University of Chinese Academy of Sciences, Beijing, China}
		\address[bsit]{CAS Center for Excellence in Brain Science and Intelligence Technology, Beijing, China}
		
		
		
		
		
		\begin{abstract}
			In sequence to sequence generation tasks (e.g. machine translation and abstractive summarization), inference is generally performed in a left-to-right manner to produce the result token by token.
			The neural approaches, such as LSTM and self-attention networks, are now able to make full use of all the predicted history hypotheses from left side during inference, but cannot meanwhile access any future (right side) information and usually generate unbalanced outputs in which left parts are much more accurate than right ones.
			In this work, we propose a synchronous bidirectional inference model to generate outputs using both left-to-right and right-to-left decoding simultaneously and interactively.
			First, we introduce a novel beam search algorithm that facilitates synchronous bidirectional decoding. Then, we present the core approach which enables left-to-right and right-to-left decoding to interact with each other, so as to utilize both the history and future predictions simultaneously during inference.
			We apply the proposed model to both LSTM and self-attention networks. In addition, we propose two strategies for parameter optimization. The extensive experiments on machine translation and abstractive summarization demonstrate that our synchronous bidirectional inference model can achieve remarkable improvements over the strong baselines.
		\end{abstract}
		
		
	\end{frontmatter}
	
	
	\section{Introduction}
	
	Many tasks in natural language processing, such as machine translation, abstractive summarization and chatbot, can be formalized as a sequence to sequence (seq2seq) generation problem which takes a sequence as input (e.g. source language sentence) and produces another sequence as output (e.g. target language translation). Generally, the seq2seq framework performs inference in a left-to-right (L2R) manner and predicts the current output token conditioned on previously generated tokens. Existing methods mainly focus on how to fully exploit the already predicted outputs on the left. And the recently proposed neural architectures for sequence generation including recurrent networks \cite{sutskever2014sequence,bahdanau2015neural}, convolutional networks \cite{gehring2017convolutional} and self-attention ones (known as Transformer) \cite{vaswani2017attention} facilitate the exploration of all the history information during inference. 
	
	\begin{table*}
		\centering
		\begin{tabular}{c|c|c|c}
			\hline
			\bf Architecture & \bf Direction & \bf First Four & \bf Last Four \\ \hline
			\multirow{2}{*}{LSTM} & L2R & \bf 36.35\% & 31.64\% \\
			{} &  R2L & 31.22\% & \bf 34.01\% \\ \hline \hline
			\multirow{2}{*}{Transformer (Self-Attention Network)} & R2L & \bf 40.21\% & 35.10\% \\
			&  R2L & 35.67\% & \bf 39.47\% \\
			\hline
		\end{tabular}
		\caption{Matching accuracy of the first and last four tokens between model predictions and references in NIST Chinese-English machine translation tasks. L2R denotes conventional left-to-right inference while R2L denotes right-to-left inference.
		} \label{match-acc}
	\end{table*}
	
	
	However, conventional seq2seq models cannot access the future predictions on the right and usually produce unbalanced outputs in which left parts are much more accurate than right ones. The phenomenon is similar for right-to-left (R2L) inference where the right parts are better. In order to have a more intuitive understanding, we have investigated both L2R and R2L inferences using LSTM \cite{hochreiter1997long} and self-attention networks (SAN) \cite{vaswani2017attention} on the typical sequence generation task, namely machine translation. Table~\ref{match-acc} shows the matching accuracy of the first and last four tokens between model predictions and references. It is obvious that left-to-right inference performs much better on predictions of head tokens while right-to-left inference excels in tail token predictions. Intuitively, it is a promising direction to combine the merits of bidirectional inferences and make full use of both history and future contexts.
	
	Researchers have made great efforts to take advantages of both L2R and R2L inferences. \cite{liu2016agreementb,zhang2018regularizing} enforce the agreement between L2R and R2L predictions during training, and then L2R inference will be improved accordingly. \cite{liu2016agreementa,wang2017sogou} employ R2L model to rerank the $n$-best hypotheses of the L2R model. \cite{zhang2018asynchronous} first obtains the R2L outputs and optimizes the L2R inference model based on both of the original input and the R2L outputs. Despite the performance improvement, these approaches suffer from two issues. On one hand, they have to train two separate seq2seq models for L2R and R2L inferences respectively. On the other hand, the two models cannot interact with each other during inference.
	
\begin{figure}
		\centering
		\includegraphics[scale=.65]{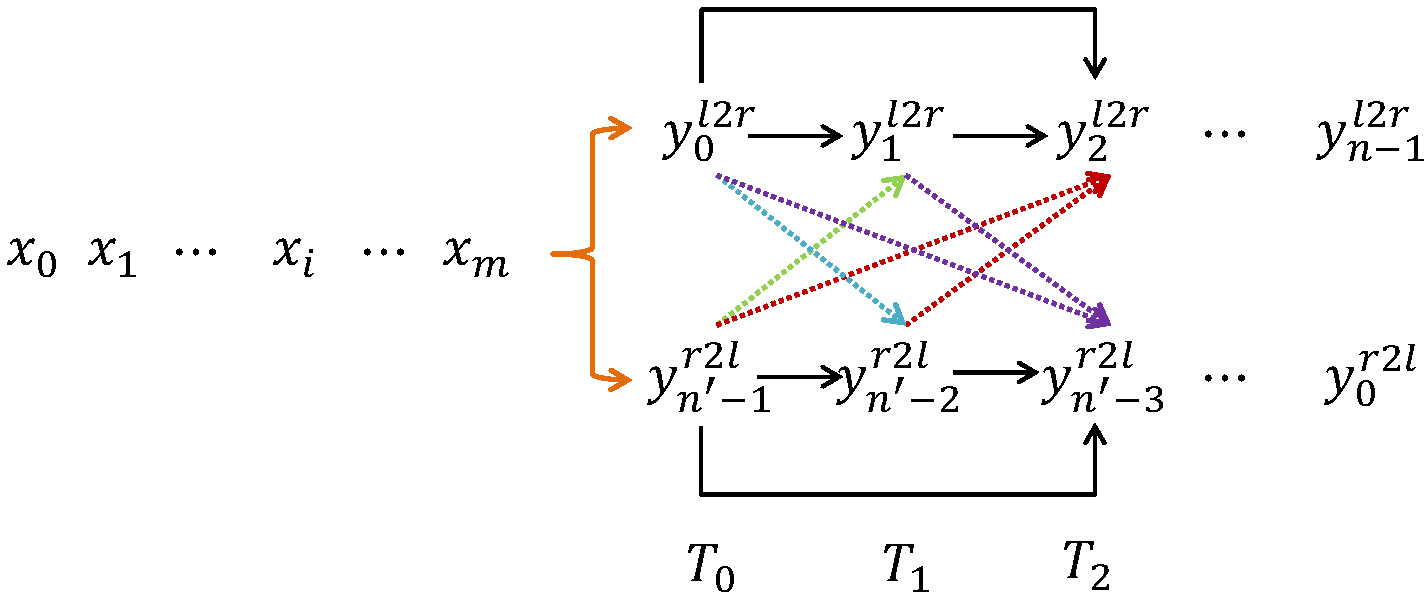}
		\caption{A simple illustration of our synchronous bidirectional inference model in which L2R and R2L models interact with each other. The left part is the input sequence and the right part denotes the scheme of synchronous bidirectional inference. Take generating $y_2^{l2r}$ as an example. We cannot only utilize the past predictions ($y_0^{l2r}$, $y_1^{l2r}$) of the L2R inference, but also could leverage the future contexts ($y_{n'-1}^{r2L}$, $y_{n'-2}^{r2l}$) which have been already predicted by the R2L inference. It is similar when predicting $y_{n'-3}^{r2l}$. Note that $y_0^{l2r}$ is not necessary the same as $y_0^{r2l}$ and $n$ may be different from $n'$. The final output sequence will be $y_0^{l2r}y_1^{l2r}y_2^{l2r}\cdots y_{n-1}^{l2r}$ if L2R inference wins. It will be $y_0^{r2l}\cdots y_{n'-3}^{r2l} y_{n'-2}^{r2l} y_{n'-1}^{r2l}$ otherwise.}
		\label{bi-infer}       
	\end{figure}

	In this article, we propose a synchronous bidirectional inference model that produces outputs using both L2R and R2L decoding simultaneously and interactively. We first introduce a novel beam search algorithm to accommodate L2R and R2L inferences at the same time. At each timestep during inference, each half beam retain the hypotheses from L2R and R2L inferences respectively and each hypothesis is generated by utilizing already predicted outputs from both directions. The interaction between L2R and R2L inferences is realized through a synchronous attention model that attempts to leverage both the history and future sequential predictions simultaneously during inference. Fig.~\ref{bi-infer} gives a simple illustration of the proposed synchronous bidirectional inference model. The middle part in color on the right of Fig.~\ref{bi-infer} is the core of our model. L2R and R2L inferences interact with each other in an implicit way illustrated by the colored part. The arrows indicate the information passing flow. Solid arrows in black show the conventional history context dependence while dotted arrows in color introduce the future context dependence on the other inference direction.  For example, besides the past predictions ($y_0^{l2r}$, $y_1^{l2r}$), L2R inference can also utilize the future contexts ($y_{n'-1}^{r2L}$, $y_{n'-2}^{r2l}$)  generated by the R2L inference when predicting $y_2^{l2r}$.
	
	As we mentioned above, there are many sequence to sequence models. To test the generalization capacity of our model, we apply the proposed synchronous bidirectional inference model into two representative seq2seq frameworks using LSTM and self-attention networks. Furthermore, we propose two optimization strategies to train network parameters. We choose machine translation and abstractive sentence summarization as the testbed to verify the effectiveness of the models. The extensive experiments demonstrate that our proposed model remarkably outperforms the strong baselines.
	
	\section{Synchronous Bidirectional Inference}
	The task of sequence to sequence learning is to find the most probable output sequence $y=y_0y_1\cdots y_{n-1}$ which maximizes the following conditional probability given the input sequence $x=x_0x_1\cdots x_{m-1}$.
	
	\begin{equation}
	P(y|x)=P(y_0y_1\cdots y_{n-1}|x_0x_1\cdots x_{m-1})
	\label{x2y}
	\end{equation}
	
	Unlike sequential labeling tasks in which $y$ shares the same length as $x$, the output length of $y$ is unknown until the inference process ends and in most cases the token numbers are different between output and input sequences in seq2seq learning tasks. For instance, the result summary (output sequence) should be much shorter than the original text (input sequence) in the summarization task.
	
	\subsection{Unidirectional Inference}
	
	Conventionally, Equation~\ref{x2y} is decomposed in a left-to-right manner as follows:
	
	\begin{eqnarray}
	P(y|x) = \prod_{i=0}^{n-1} p(y_i|y_0\cdots y_{i-1}, x)
	\label{l2rdecoposition}
	\end{eqnarray}
	Since the search space $V$ ($y_i \in V$) is very large and contains tens of thousands of entries in most cases, a beam search algorithm is usually employed to approximately find the most probable output sequence according to the history predictions $y_0\cdots y_{i-1}$ and the input sequence $x$. Currently, neural methods such as LSTM, conventional networks and self-attention ones can model the conditional probability $p(y_i|y_0\cdots y_{i-1}, x)$ more and more accurately, but leaves the future contexts unexplored.
	
	In order to leverage the right hand information, Equation~\ref{x2y} can also be decomposed in a right-to-left manner as follows:
	
	\begin{eqnarray}
	P(y|x) = \prod_{i=n-1}^{0} p(y_i|y_{i+1}\cdots y_{n-1}, x)
	\label{r2ldecomposition}
	\end{eqnarray}
	Using this decomposition, the right-side hypotheses $y_{i+1}\cdots y_{n-1}$ are available when predicting $y_i$, while the left-side predictions are still missing.

	\subsection{Synchronous Bidirectional Beam Search}
	
	\begin{algorithm}
		\caption{Synchronous Bidirectional Beam Search Algorithm}
		\KwIn{Input sequence $x=\langle s \rangle x_0\cdots x_j \cdots x_{m-1} \langle /s \rangle$, beam size $K$, maximum length of output sequence $MaxLen$}
		\KwOut{Optimal output sequence $y=y_0\cdots y_i \cdots y_{n-1}$}
		Initialize complete hypothesis list $B=\Phi$, left-to-right temporary hypothesis list $B_{l-tmp}=\Phi$ and partial hypothesis list $B_{l-part}^0=\{\langle l2r \rangle\}$, right-to-left temporary hypothesis list $B_{r-tmp}=\Phi$ and partial hypothesis list $B_{r-part}^0=\{\langle r2l \rangle\}$; //$\langle l2r \rangle$ and $\langle r2l \rangle$ are tags indicating inference direction\;
		
		\For{$i=1;i \le MaxLen$}
		{
			$B_{l-tmp}$=ExpandHypo$(B_{l-part}^{i-1}, B_{r-part}^{i-1})$\;
			$B_{r-tmp}$=ExpandHypo$(B_{r-part}^{i-1}, B_{l-part}^{i-1})$\;
			$[B_{l-part}^{i},B]$=UpdateHypo$(B_{l-tmp},B_{l-part}^{i},B)$\;
			$[B_{r-part}^{i},B]$=UpdateHypo$(B_{r-tmp},B_{l-part}^{i},B)$\;
			$B_{l-tmp}=\Phi$\;
			$B_{r-tmp}=\Phi$\;
		}
		\eIf{$B \ne \Phi$}
		{
			sort $B$ in decending order\;
			$y=B[0]$\;
		}
		{
			\eIf{$p(B_{l-part}^{MaxLen}[0]) \ge p(B_{r-part}^{MaxLen}[0])$}
			{
				$y=B_{l-part}^{MaxLen}[0]$\;
			}
			{
				$y=B_{r-part}^{MaxLen}[0]$\;
			}
		}
		\eIf{$y[0]=\langle l2r \rangle$}
		{
			return $y$\;
		}
		{
			return reversed $y$\;
		}
	\end{algorithm}
	
	Ideally, we expect to utilize both the past and future contexts ($y_0\cdots y_{i-1}$ and $y_{i+1}\cdots y_{n-1}$) when determining the best prediction of $y_i$. However, it is contradictory to some extent. Predicting $y_i$ needs $y_{i+1}$ on the right, while determining $y_{i+1}$ requires $y_{i}$ on the left. Obviously, it is impractical to use the whole contexts of both sides ($y_0\cdots y_{i-1}$ and $y_{i+1}\cdots y_{n-1}$) in a single inference model. We take a step back and attempt to explore the bidirectional contexts as many as possible if not all.
	
	We propose a synchronous bidirectional inference model in which left-to-right and right-to-left inferences perform in parallel while keeping interaction with each other. In this way, Equation~\ref{x2y} is decomposed as follows:
	
	\begin{equation}
	P(y|x) =
	\begin{cases}
	\prod_{i=0}^{n-1} p(\overrightarrow{y_i}|\overrightarrow{y}_0\cdots \overrightarrow{y}_{i-1}, x, \overleftarrow{y}_0\cdots \overleftarrow{y}_{i-1}) & \text{if L2R} \\
	\prod_{i=0}^{n'-1} p(\overleftarrow{y}_i|\overleftarrow{y}_0\cdots \overleftarrow{y}_{i-1}, x, \overrightarrow{y}_0\cdots \overrightarrow{y}_{i-1}) & \text{if R2L}
	\end{cases}
	\label{bidecoposition}
	\end{equation}
	Equation~\ref{bidecoposition} says that the bidirectional inference model accommodates L2R and R2L decoding at the same time. At timestep $i$, we have already generated the left $i-1$ hypotheses $\overrightarrow{y}_0\cdots \overrightarrow{y}_{i-1}$ with L2R inference and the right $i-1$ predictions $\overleftarrow{y}_0\cdots \overleftarrow{y}_{i-1}$ with R2L inference. Thus, different from equation~\ref{l2rdecoposition} and equation~\ref{r2ldecomposition}, both-side predictions can be utilized as contexts in the above bidirectional composition.
	
	{\bf Algorithm 1} shows the beam search procedure of the synchronous bidirectional inference model. The working flow is similar to the unidirectional beam search. We keep three kinds of lists. $B$ is employed to store complete hypotheses. $(B_{l-tmp}, B_{r-tmp})$ and $(B_{l-part}, B_{r-part})$ are used to maintain the temporary and partial hypotheses at each decoding timestep for L2R and R2L inferences respectively. Lines 2-9 in {\bf Algorithm 1} is the main part of the beam search algorithm. At timestep $i$, L2R and R2L inferences perform in parallel but interactively to expand the partial hypotheses $B_{l-part}^{i-1}$ and $B_{r-part}^{i-1}$ from the previous timestep (lines 3-4). Then, the complete hypothesis list $B$ and the partial hypothesis list $B_{l-part}^i, B_{r-part}^i$ will be updated according to the temporary hypothesis list $(B_{l-tmp}^{i-1}, B_{r-tmp}^{i-1})$ (lines 5-6).
	
	{\bf Algorithm 2} and {\bf Algorithm 3} respectively detail the hypothesis expansion procedure and hypothesis update process. {\bf Algorithm 3} is trivial and is the same as the conventional unidirectional beam search. The algorithm {\em ExpandHypo}$(B_{f-part}^{i-1},B_{b-part}^{i-1})$ is the key for synchronous bidirectional inference ({\bf SBInfer}). In {\bf Algorithm 2}, for a partial hypothesis in $B_{f-part}^{i-1}$, we calculate the probability of each candidate token in the target vocabulary by utilizing both of the history context $B_{f-part}^{i-1}$ and the future information $B_{b-part}^{i-1}$ through the function {\em SBInfer}$(cand, B_{f-part}^{i-1}, B_{b-part}^{i-1})$.
	
	Obviously, the function {\em SBInfer}$(cand, B_{f-part}^{i-1}, B_{b-part}^{i-1})$ is the most important part and requires specific design for different seq2seq architectures. Next, we introduce how to implement the function {\em SBInfer}$(cand, B_{f-part}^{i-1}, B_{b-part}^{i-1})$ for both LSTM-based and self-attention based seq2seq networks.

	\begin{algorithm}
		\caption{ExpandHypo$(B_{f-part}^{i-1}, B_{b-part}^{i-1})$.}
		\KwIn{Partial hypothesis list $B_{f-part}^{i-1}$ for current decoding direction and $B_{b-part}^{i-1}$ for the opposite decoding direction}
		\KwOut{temporary hypothesis list $B_{tmp}=\Phi$}
		
		\For{$cand$ in $B_{f-part}^{i-1}$}
		{
			\For{$y_i^{*}$ in $TargetVocab$}
			{
				$cand=cand + y_i^{*}$\;
				$p(cand)$=SBInfer$(cand, B_{f-part}^{i-1}, B_{b-part}^{i-1})$ \;
				$B_{tmp}=B_{tmp} \bigcup \{ cand \}$\;
			}
		}
		
		sort $B_{tmp}$ in a decending order\;
		$B_{tmp}=B_{tmp}[0:K/2]$\;
		return $B_{tmp}$\;
	\end{algorithm}
	
	\begin{algorithm}
		\caption{UpdateHypo$(B_{tmp}, B_{part}, B)$}
		\KwIn{Temporary hypothesis list $B_{tmp}=\Phi$, partial hypothesis list $B_{part}$, and complete hypothesis list $B$}
		\KwOut{Partial hypothesis list $B_{part}$ and complete hypothesis list $B$}
		
		\For{$cand$ in $B_{tmp}$}
		{
			\eIf{$\langle /s \rangle = cand[-1]$}
			{
				$B = B \bigcup \{cand\}$\;
				\If{$K \ge |B|$}
				{
					break\;
				}
			}
			{
				$B_{part} = B_{part} \bigcup \{ cand \}$\;
			}
		}
		return $[B_{part},B]$\;
		
	\end{algorithm}
	
	\section{Synchronous Bidirectional Inference for LSTM-based Seq2Seq Framework}
	
	No matter what kind of network architecture is used, all Seq2Seq frameworks consist of an encoder and a decoder. Given an input sequence $x=(x_0,x_2,\cdots,x_{m-1})$, the encoder transforms $x$ into a sequence of abstract context representations $C=(\mathbf{h}_0,\mathbf{h}_1,\cdots,\mathbf{h}_{m-1})$ whose size is the same as the length of the input text. Then, from the context vectors $C$ the decoder generates the output sequence $y=(\overrightarrow{y}_0,\overrightarrow{y}_1,\cdots,\overrightarrow{y}_{n-1})$ one token each time by maximizing the probability of $p(\overrightarrow{y}_i|\overrightarrow{y}_{<i},C)$ with a left-to-right inference model.
	
	Hereafter, we leverage $\mathbf{x}_j$  and $\mathbf{y}_i$ to denote the word embeddings corresponding to the input and output tokens $x_j$ and $y_i$. Next, we briefly review the encoder introducing how to obtain $C$ and the decoder addressing how to calculate $p(\overrightarrow{y}_i|\overrightarrow{y}_{<i},C)$ for the conventional LSTM-based Seq2Seq architecture. Then, we propose to enable synchronous bidirectional inference {\em SBInfer}$(cand, B_{f-part}^{i-1}, B_{b-part}^{i-1})$ in the LSTM-based architecture.
	
	\subsection{LSTM-based Seq2Seq Framework}
	
	The {\bf encoder} employs $L$ stacked LSTM layers to learn the context vectors $C=(\mathbf{h}_0,\mathbf{h}_1,\cdots,\mathbf{h}_{m-1})$. In the $l$-th layer ($l>1$), $\mathbf{h}^l_j$ is calculated as follows:
	
	\begin{equation}
	\mathbf{h}^l_j = LSTM(\mathbf{h}^l_{j-1}, \mathbf{h}^{l-1}_j)
	\end{equation}
	
	In the first layer ($l=1$), $\mathbf{h}^1_j $ is obtained through a bidirectional LSTM:
	
	\begin{equation}
	\overrightarrow{\mathbf{h}}^1_j = LSTM(\overrightarrow{\mathbf{h}}^1_{j-1}, x_j)
	\end{equation}
	
	\begin{equation}
	\overleftarrow{\mathbf{h}}^1_j = LSTM(\overleftarrow{\mathbf{h}}^1_{j+1}, x_j)
	\end{equation}
	
	Given $\overrightarrow{\mathbf{h}}^1_j$ and $\overleftarrow{\mathbf{h}}^1_j$, $\mathbf{h}^1_j $ is calculated with a feed-forward neural network $\mathbf{h}^1_j=tanh(W^l_h\cdot\overrightarrow{\mathbf{h}}^1_j + W^r_h\cdot \overleftarrow{\mathbf{h}}^1_j + b_h)$.
	
	The {\bf decoder} computes the conditional probability $p(\overrightarrow{y}_i|\overrightarrow{y}_{<i},C)$ with the help of {\bf attention} mechanism \cite{bahdanau2015neural} that leverages different input context $\mathbf{c}_i$ at different decoding time step:
	
	\begin{equation}
	p(\overrightarrow{y}_i|\overrightarrow{y}_{<i},C) = p(\overrightarrow{y}_i|\overrightarrow{y}_{<i},\mathbf{c}_i) = softmax(W\overrightarrow{\mathbf{z}}_i)
	\label{l2rprob}
	\end{equation}
	where $\overrightarrow{\mathbf{z}}_i$ is the attention output:
	
	\begin{equation}
	\overrightarrow{\mathbf{z}}_i = tanh(W_c[\overrightarrow{\mathbf{z}}^L_i;\mathbf{c}_i])
	\label{leftctx}
	\end{equation}
	in which $\overrightarrow{\mathbf{z}}^L_i$ is the top hidden state of the decoder network and $\overrightarrow{\mathbf{z}}^l_i$ in the $l$-th layer is computed using the following formula:
	
	\begin{equation}
	\overrightarrow{\mathbf{z}}^l_i = LSTM(\overrightarrow{\mathbf{z}}^{l}_{i-1}, \overrightarrow{\mathbf{z}}^{l-1}_i)
	\end{equation}
	
	If $l=1$, $\overrightarrow{\mathbf{z}}^1_i$ will be calculated by combining $\overrightarrow{\mathbf{z}}_{i-1}$ as feed input \cite{luong2015effective}:
	
	\begin{equation}
	\overrightarrow{\mathbf{z}}^1_i = LSTM(\overrightarrow{\mathbf{z}}^1_{i-1}, y_{i-1}, \overrightarrow{\mathbf{z}}_{i-1})
	\end{equation}
	
	The dynamic context $\mathbf{c}_i$ is the weighted sum of the source-side context vectors and is calculated by the attention model:
	
	\begin{equation}
	\mathbf{c}_i = \sum_{j=0}^{m-1} \alpha_{ij}\mathbf{h}_{j}
	\end{equation}
	
	where $\alpha_{ij}$ is a normalized item calculated as follows:
	
	\begin{equation}
	e_{ij} =v^{\intercal}_a tanh(W_a\overrightarrow{\mathbf{z}}^L_i+U_a{\mathbf{h}_j})
	\label{align:1}
	\end{equation}
	
	\begin{equation}
	\alpha_{ij} = \frac{exp(e_{ij})}{{\sum_{j'} exp(e_{ij'})}}
	\label{att:1}
	\end{equation}
	
\begin{figure}[!t]
		\centering
		\includegraphics[scale=.5]{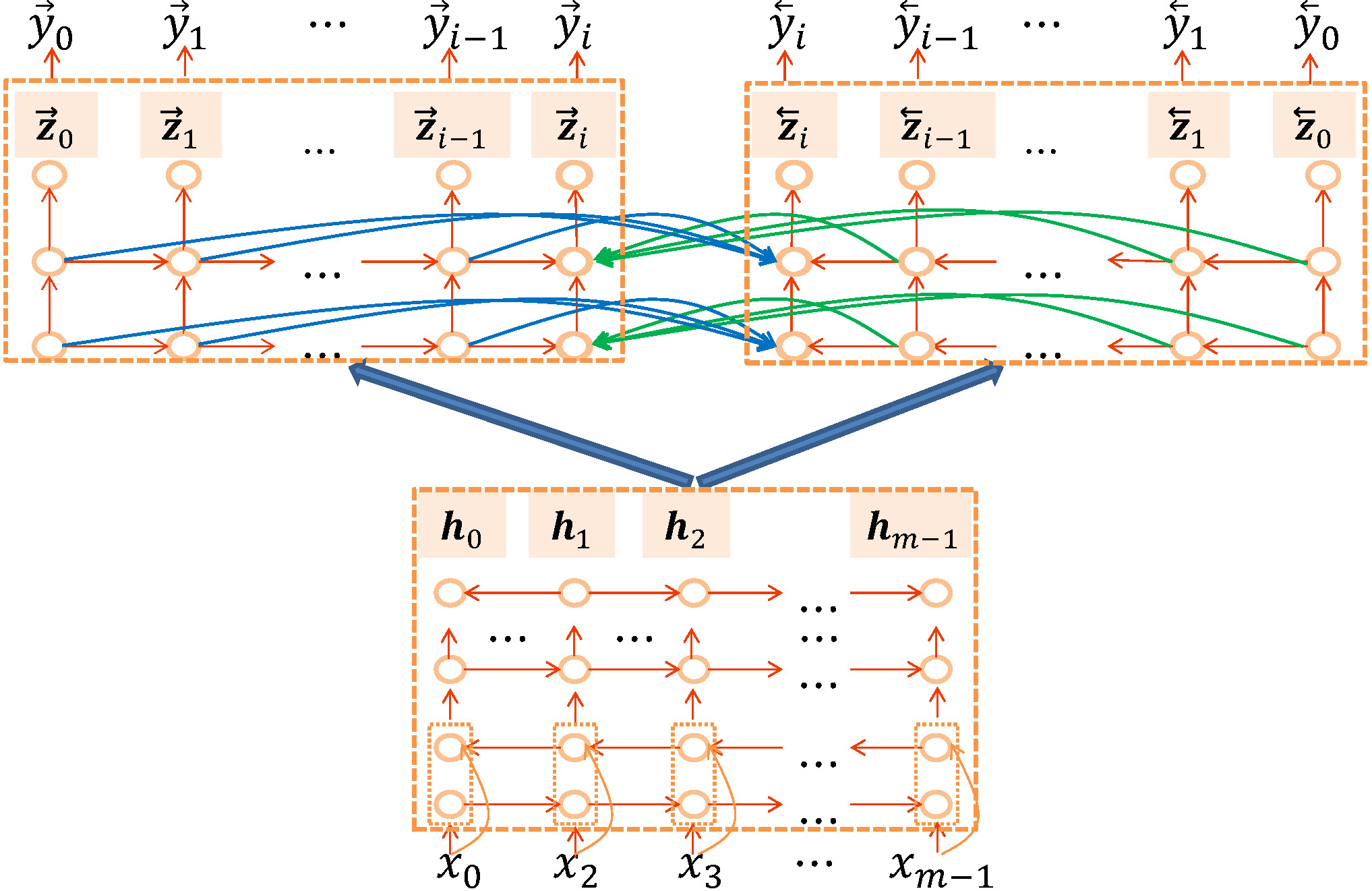}
		\caption{Bidirectional inference model for seq2seq framework with LSTM architecture.}
		\label{fig:01}       
	\end{figure}

	The greater the value of the variable $\alpha_{ij}$, the more contribution of the $j$-th input token to the generation of the $i$-th output word. The left part in Fig.~\ref{fig:01} gives the overall illustration of this unidirectional inference model for LSTM-based Seq2Seq framework. Note that residual connections and layer normalizations are employed as well and they are neglected in the description for simplicity.
	
	\subsection{Synchronous Bidirectional Inference for LSTM-based Architecture}
	
	In synchronous bidirectional inference, $p(\overrightarrow{y}_i)$ is calculated with both history and future contexts according to Equation~\ref{bidecoposition}: $p(\overrightarrow{y_i}|\overrightarrow{y}_0\cdots \overrightarrow{y}_{i-1}, x, \overleftarrow{y}_0\cdots \overleftarrow{y}_{i-1})$. The previous section introduces the way to use input $x$ and history contexts $\overrightarrow{y}_0\cdots \overrightarrow{y}_{i-1}$ in Equation~\ref{l2rprob}: $p(\overrightarrow{y_i}|\overrightarrow{y}_{<i}, C)$. The synchronous bidirectional inference adopts the same mechanism as follows:
	
	\begin{equation}
	p(\overrightarrow{y}_i|\overrightarrow{y}_{<i},C,\overleftarrow{y}_{<i}) = p(\overrightarrow{y}_i|\overrightarrow{y}_{<i},\mathbf{c}_i, \overleftarrow{y}_{<i}) = softmax(W\overrightarrow{\mathbf{z}}_i)
	\label{sbiprob}
	\end{equation}
	
	Different from unidirectional inference, the synchronous bidirectional inference calculates the attention output $\overrightarrow{\mathbf{z}}_i$ with both L2R and R2L predictions:
	
	\begin{equation}
	\overrightarrow{\mathbf{z}}_i = tanh(W_c[\overrightarrow{\mathbf{z}}^L_i;\mathbf{c}_i;\overleftarrow{\mathbf{cz}}_i])
	\label{sbictx}
	\end{equation}
	where the future context $\overleftarrow{\mathbf{cz}}_i$ is obtained using another attention model as illustrated with green arrows in Fig.~\ref{fig:01}.
	
	\begin{equation}
	\overleftarrow{\mathbf{cz}}_i = \sum_{k=0}^{i-1} \overleftarrow{\gamma}_{ik}\overleftarrow{\mathbf{z}}^L_k
	\end{equation}
	in which $\overleftarrow{\gamma}_{ik}$ is a normalized coefficient:
	
	\begin{equation}
	e_{ik} =v^{\intercal}_z tanh(W_z\overrightarrow{\mathbf{z}}^L_i+U_z{\overleftarrow{\mathbf{z}}^L_k})
	\label{align:2}
	\end{equation}
	
	\begin{equation}
	\overleftarrow{\gamma}_{ik} = \frac{exp(e_{ik})}{{\sum_{k'} exp(e_{ik'})}}
	\label{att:2}
	\end{equation}
	
	It should be noted that L2R and R2L inferences perform simultaneously in parallel. Thus, when calculating $p(\overrightarrow{y}_i|\overrightarrow{y}_{<i},C,\overleftarrow{y}_{<i})$, we can as well compute $p(\overleftarrow{y}_i|\overleftarrow{y}_{<i},C,\overrightarrow{y}_{<i})$ at the same time in a similar way.
	
	\begin{equation}
	p(\overleftarrow{y}_i|\overleftarrow{y}_{<i},C,\overrightarrow{y}_{<i}) = p(\overleftarrow{y}_i|\overleftarrow{y}_{<i},\mathbf{c}_i, \overrightarrow{y}_{<i}) = softmax(W\overleftarrow{\mathbf{z}}_i)
	\label{sbiprob-r2l}
	\end{equation}
	
	\begin{equation}
	\overleftarrow{\mathbf{z}}_i = tanh(W_c[\overleftarrow{\mathbf{z}}^L_i;\mathbf{c}_i;\overrightarrow{\mathbf{cz}}_i])
	\label{sbictx-r2l}
	\end{equation}
	Where the left context $\overrightarrow{\mathbf{cz}}_i$ is obtained using a similar attention model as illustrated with blue arrows in Fig.~\ref{fig:01}.
	
	\begin{equation}
	\overrightarrow{\mathbf{cz}}_i = \sum_{k=0}^{i-1} \overrightarrow{\gamma}_{ik}\overrightarrow{\mathbf{z}}^L_k
	\end{equation}

	\section{Synchronous Bidirectional Inference for Self-attention based Framework}
	
	The self-attention based Seq2Seq framework is known as {\bf Transformer} \cite{vaswani2017attention}. In this section, we first give an overview of Transformer and then propose the implementation of synchronous bidirectional inference {\em SBInfer}$(cand, B_{f-part}^{i-1}, B_{b-part}^{i-1})$ in Transformer.
	
	\subsection{Transformer}
	The Transformer also follows the encoder-decoder architecture. The encoder includes $L$ identical layers and each layer is composed of two sub-layers: the self-attention sub-layer followed by the feed-forward sub-layer.
	
	The decoder also consists of $L$ identical layers. Each layer has three sub-layers. The first one is the masked self-attention mechanism. The second one is the decoder-encoder attention sub-layer and the third one is the feed-forward sub-layer{\footnote{Residual connection and layer normalization are performed for each sub-layer in both encoder and decoder.}}.
	
	Obviously, the key component is the attention mechanism{\footnote{In fact, multi-head attention is employed and we just introduce basic attention for simplicity.}}. The three kinds of attention mechanisms can be formalized into the same formula.
	
	\begin{equation}
	Attention(q,K,V)=softmax(\frac{qK^T}{\sqrt{d_k}})V
	\label{transformerattention}
	\end{equation}
	Where $q$, $K$ and $V$ stand for a query, the key list and the value list respectively. $d_k$ is the dimension of the key.
	
	For the self-attention in encoder, the queries, keys and values are from the same layer. For example, if we calculate the output of the first layer in the encoder at the $j$-th position. The query is vector $\mathbf{x}_j${\footnote{Suppose $\mathbf{x}_j$ is the sum vector of input token embedding and the positional embedding.}}. The keys and values are the same and both are the embedding matrix $\mathbf{x}=[\mathbf{x}_0 \cdots \mathbf{x}_{m-1}]$. Using Equation~\ref{transformerattention} followed by a feed-forward network, we can get the representation of the second layer. After $L$ layers, we obtain the input contexts $C=[\mathbf{h}_0, \cdots, \mathbf{h}_{m-1}]$.
	
	The masked self-attention in decoder is similar to that of encoder except that the query at the $i$-th position can only attend to positions before $i$, since the predictions after $i$-th position are not available in the auto-regressive unidirectional inference.
	
	\begin{equation}
	\overrightarrow{\mathbf{z}}^{past}_i=Attention(\overrightarrow{q}_i,\overrightarrow{K}_{\le i},\overrightarrow{V}_{\le i})=softmax(\frac{\overrightarrow{q}_i\overrightarrow{K}^T_{\le i}}{\sqrt{d_k}})\overrightarrow{V}_{\le i}
	\label{history-attention}
	\end{equation}
	
	The decoder-encoder attention mechanism is the same as that of LSTM-based Seq2Seq architecture. The query is the output of the masked self-attention sub-layer $\overrightarrow{\mathbf{z}}^{past}_i$. The keys and values are the same encoder contexts $C$. The feed-forward sub-layer is then applied to yield the output of a whole layer. After $L$ such layers, we obtain the final hidden state $\overrightarrow{\mathbf{z}}_i$s. Softmax function (Equation~\ref{l2rprob}) is then employed to predict the output $\overrightarrow{y}_i$. Left part in Fig.~\ref{fig:02} depicts the overall architecture of Transformer.

	\subsection{Synchronous Bidirectional Inference for Transformer}
	In synchronous bidirectional inference, the essential difference lies in the improvement over the masked self-attention mechanism for decoder. In standard Transformer, the masked self-attention model calculates the output $\overrightarrow{\mathbf{z}}_i$ ($\overrightarrow{\mathbf{z}}^{past}_i$) using only the history contexts. In contrast, synchronous bidirectional inference performs L2R and R2L decoding in parallel and interactively. At the $i$-th timestep, L2R and R2L inferences have already generated $i-1$ outputs $\overrightarrow{\mathbf{z}}^{past}_{\le i}=(\overrightarrow{\mathbf{z}}^{past}_0 \cdots \overrightarrow{\mathbf{z}}^{past}_{i-1})$ and $\overleftarrow{\mathbf{z}}^{past}_{\le i}=(\overleftarrow{\mathbf{z}}^{past}_0 \cdots \overleftarrow{\mathbf{z}}^{past}_{i-1})$. Therefore, both $\overrightarrow{\mathbf{z}}^{past}_{\le i}$ and $\overleftarrow{\mathbf{z}}^{past}_{\le i}$ can be employed to compute $\overrightarrow{\mathbf{z}}_i$.
	
\begin{figure}[!t]
		\centering
		\includegraphics[scale=.5]{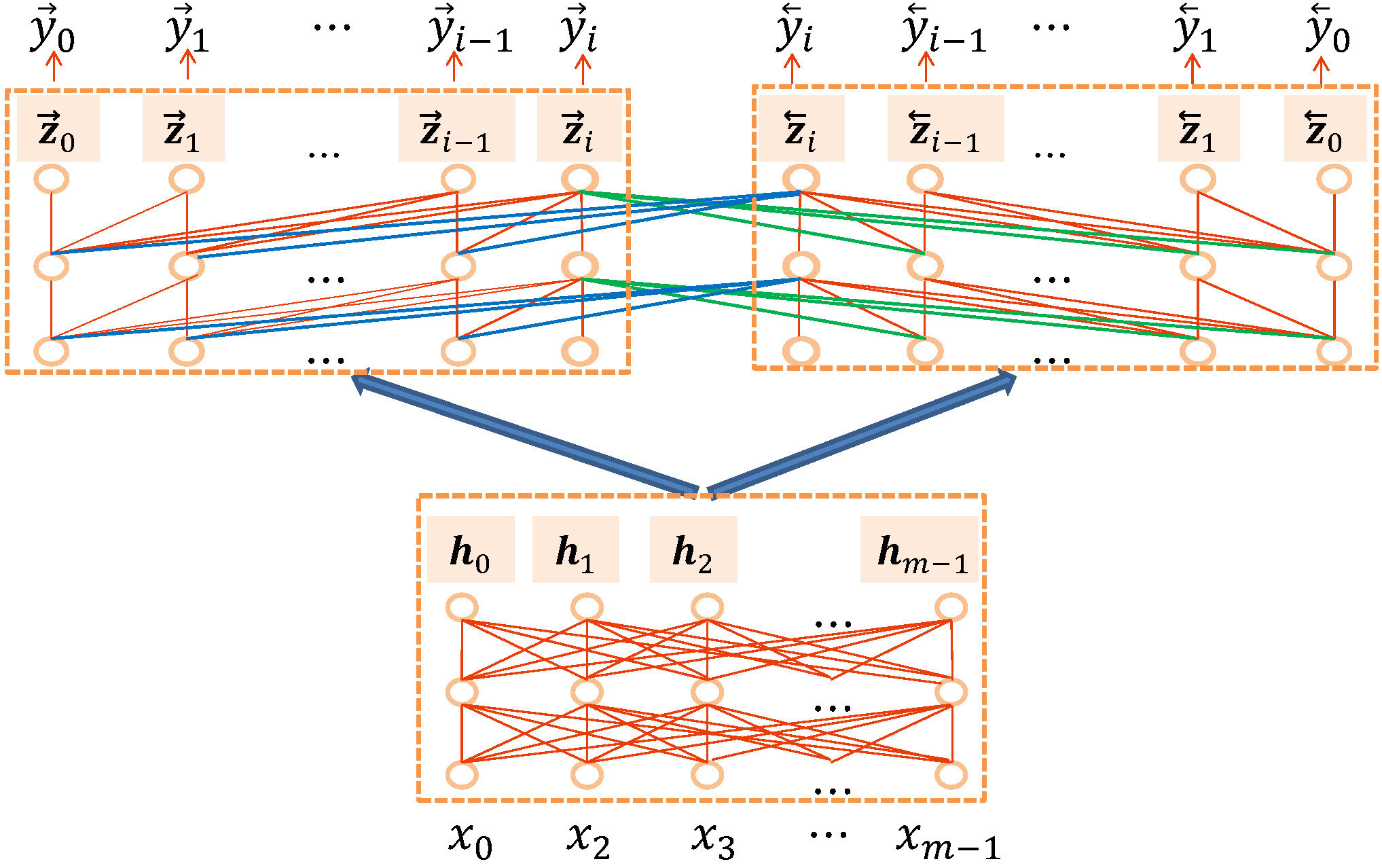}
		\caption{Bidirectional inference model for seq2seq framework with self-attention architecture.}
		\label{fig:02}       
	\end{figure}

	Accordingly, we design two self-attention mechanisms to handle history contexts $\overrightarrow{\mathbf{z}}^{past}_{\le i}$ and future contexts $\overleftarrow{\mathbf{z}}^{past}_{\le i}$ respectively. In addition to Equation~\ref{history-attention} that utilizes history information, we propose another self-attention mechanism to leverage the future information generated by the opposite inference direction.

	\begin{equation}
	\overrightarrow{\mathbf{z}}^{future}_i = Attention(\overrightarrow{q}_i,\overleftarrow{K}_{\le i},\overleftarrow{V}_{\le i})=softmax(\frac{\overrightarrow{q}_i\overleftarrow{K}^T_{\le i}}{\sqrt{d_k}})\overleftarrow{V}_{\le i}
	\label{future-attention}
	\end{equation}
	where $\overrightarrow{q}_i$ is the query (i.e. the embedding of the $(i-1)$-th output $\overrightarrow{y}_{i-1}$) from the L2R decoder. $\overleftarrow{K}_{\le i}$ and $\overleftarrow{V}_{\le i}$ are keys and values (i.e. the embeddings of the previous $i-1$ predictions $\overleftarrow{y}_0, \cdots, \overleftarrow{y}_{i-1}$) from the R2L decoder. Fig.~\ref{fig:02} illustrates how synchronous bidirectional inference performs. When producing $\overrightarrow{z}_i$, the orange lines denote the original masked self-attention with history while the green lines indicate the self-attention with future contexts.
	
	Finally, we introduce a function to combine $\overrightarrow{\mathbf{z}}^{past}_i$ and  $\overrightarrow{\mathbf{z}}^{future}_i $ to obtain a new representation $\overrightarrow{z}_i$ that encodes both past and future contexts.
	
	\begin{equation}
	\overrightarrow{\mathbf{z}}_i = f(\overrightarrow{\mathbf{z}}^{past}_i, \overrightarrow{\mathbf{z}}^{future}_i)=\overrightarrow{\mathbf{z}}^{past}_i + \lambda \times tanh(\overrightarrow{\mathbf{z}}^{future}_i)
	\label{history-future-comb}
	\end{equation}
	
	For R2L decoding, $\overleftarrow{\mathbf{z}}_i $ can be calculated similarly in parallel.
	
	\begin{equation}
	\overleftarrow{\mathbf{z}}^{future}_i = Attention(\overleftarrow{q}_i,\overrightarrow{K}_{\le i},\overrightarrow{V}_{\le i})=softmax(\frac{\overleftarrow{q}_i\overrightarrow{K}^T_{\le i}}{\sqrt{d_k}})\overrightarrow{V}_{\le i}
	\label{future-attention}
	\end{equation}
	
	\begin{equation}
	\overleftarrow{\mathbf{z}}_i = \overleftarrow{\mathbf{z}}^{past}_i + \lambda \times tanh(\overleftarrow{\mathbf{z}}^{future}_i)
	\end{equation}
	
	\section{Training}
	Since our synchronous bidirectional inference performs L2R and R2L decoding in parallel, L2R decoder aims to generate the gold reference $\overrightarrow{y}$ during training while R2L decoder attempts to produce the reversed gold reference $\overleftarrow{y}$ at the same time. Given the training data consisting of $T$ sentence pairs $(x^{(t)}, y^{(t)})_{t=1}^T$, the objective is to maximize the log-likelihood over the training data.
	
	\begin{equation}
	J(\theta)=\sum_{t=1}^{T}\big(log p(\overrightarrow{y}^{(t)}|x^{(t)}) + log p(\overleftarrow{y}^{(t)}|x^{(t)})\big)
	\end{equation}
	
	When calculating $p(\overrightarrow{y}_i^{(t)})$, L2R decoder usually employs the gold reference $\overrightarrow{y}_{<i}^{(t)}$ as the condition $p(\overrightarrow{y}_i^{(t)}|\overrightarrow{y}_{<i }^{(t)},x^{(t)})$. In synchronous bidirectional inference, a problem will arise if we directly utilize the gold reference $\overleftarrow{y}_{<i}^{(t)}$ from the other side to compute $p(\overrightarrow{y}_i^{(t)}|\overrightarrow{y}_{<i}^{(t)},\overleftarrow{y}_{<i}^{(t)},x^{(t)})$. For example, in the calculation of $p(\overrightarrow{y}_{n-1}^{(t)}|\overrightarrow{y}_{<n-1}^{(t)},\overleftarrow{y}_{<n-1}^{(t)},x^{(t)})$, $\overleftarrow{y}_{<n-1}^{(t)}$ includes $\overleftarrow{y}_{0}^{(t)}=\overrightarrow{y}_{n-1}^{(t)}$. It indicates that $\overrightarrow{y}_{n-1}^{(t)}$ is used to predict itself. Obviously, it is not reasonable.
	To address this issue during training, we propose two optimization strategies to learn network parameters.
	
	\subsection{Two-pass Training}
	In the first training pass, we learn independent L2R and R2L inference models on the training data. Then, L2R and R2L models are employed to decode the input sentences of the training data, resulting in $(x^{(t)}, \overrightarrow{y^*}^{(t)})_{t=1}^T$ and $(x^{(t)}, \overleftarrow{y^*}^{(t)})_{t=1}^T$ respectively. 
	During the second training pass, $p(\overrightarrow{y}_i^{(t)})$ is calculated using $p(\overrightarrow{y}_i^{(t)}|\overrightarrow{y}_{<i}^{(t)},\overleftarrow{y^*}_{<i}^{(t)},x^{(t)})$, indicating that the future context is the model predictions $\overleftarrow{y^*}_{<i}^{(t)}$ rather than gold reference $\overleftarrow{y}_{<i}^{(t)}$. Similarly, we calculate $p(\overleftarrow{y}_i^{(t)})$ using $p(\overleftarrow{y}_i^{(t)}|\overleftarrow{y}_{<i}^{(t)},\overrightarrow{y^*}_{<i}^{(t)},x^{(t)})$.
	
	\subsection{Fine-tuning Strategy}
	In the find-tuning strategy, we first train the parallel inference for L2R and R2L without interaction just as Equation~\ref{nointeraction} shows. Each training instance for this step is a triple $\langle x, \overrightarrow{y}, \overleftarrow{y} \rangle$.
	
	\begin{equation}
	P(y|x) =
	\begin{cases}
	\prod_{i=0}^{n-1} p(\overrightarrow{y_i}|\overrightarrow{y}_0\cdots \overrightarrow{y}_{i-1}, x) & \text{if L2R} \\
	\prod_{i=0}^{n-1} p(\overleftarrow{y}_i|\overleftarrow{y}_0\cdots \overleftarrow{y}_{i-1}, x) & \text{if R2L}
	\end{cases}
	\label{nointeraction}
	\end{equation}
	
	After this simple training procedure converges, we use this model to decode a small subset of the source sentences in the original training data (e.g. 10\% of the dataset) and get the new triple $\langle x, \overrightarrow{y^*}, \overleftarrow{y^*} \rangle$. Then, we can fine-tune our synchronous bidirectional inference model similar to the second pass of the two-pass training strategy.
	
	Compared to the two-pass training strategy, the fine-tuning strategy is much cheaper to implement since there is no need to train two separate models, to decode the entire training set and to do the second training over the whole dataset. In the experiments, we mainly employ the two-pass strategy and compare these two strategies in the experimental analysis part.
	
	\section{Experimental Setup}
	
	In our experiments, two typical seq2seq tasks of machine translation and abstractive summarization are employed to test the effectiveness of our synchronous bidirectional inference model.
	
	\subsection{Machine Translation}
	
	\subsubsection{Dataset}
	
	We evaluate the proposed synchronous bidirectional inference model on both Chinese-to-English and English-to-German translation tasks. For the Chinese-to-English task, the training data consists of about 2.1M sentence pairs extracted from LDC corpora{\footnote{LDC2000T50, LDC2002L27, LDC2002T01, LDC2002E18, LDC2003E07, LDC2003E14, LDC2003T17, LDC2004T07.}}. We choose NIST 2002 (MT02) dataset for validation. For testing, we employ NIST 2003-2006 (MT03-06) datasets. We apply Byte-Pair Encoding (BPE)\cite{sennrich2016neural} with 30K merge operations and maintain the source and target vocabularies to the most frequent 30K tokens..
	
	For the English-to-German task, we utilize the same subset of the WMT 2014 training corpus employed by \cite{vaswani2017attention,luong2015effective,shen2016minimum,zhou2016deep}. It contains 4.5M sentence pairs\footnote{All preprocessed dataset and vocabulary can be directly download in tensor2tensor website \url{https://drive.google.com/open?id=0B_bZck-ksdkpM25jRUN2X2UxMm8}.}. The
	concatenation of news-test 2012 and news-test 2013 is used as the validation set. The news-test 2014 is employed as the test set.
	
	\subsubsection{Training and Evaluation Details}
	
	For our synchronous bidirectional inference model with {\bf LSTM-based architecture}, we implement the system by reusing and modifying the open source toolkit Zoph\_RNN{\footnote{https://github.com/isi-nlp/Zoph\_RNN}} which is written in C++/CUDA and provides efficient training across multiple GPUs. The encoder includes two stacked LSTM layers and the first layer employs the bidirectional LSTMs. The decoder also contains two stacked LSTM layers followed by the softmax layer. The dimension of word embedding and the size of hidden layers are all set to 1000. The dropout rate is set to 0.2. At test time, we employ beam search with beam size $k=4$.
	
	For the synchronous bidirectional inference model with {\bf Transformer},  we modify the tensor2tensor{\footnote{https://github.com/tensorflow/tensor2tensor.}} toolkit for training and evaluation.
	We employ the Adam optimizer with $\beta_1$=0.9, $\beta_2$=0.998, and $\epsilon$=$10^{-9}$. The warmup and decay strategy for learning rate are the same as \cite{vaswani2017attention}, with 16,000 warmup steps.
	During training, we employ label smoothing of value $\epsilon_{ls}$=0.1.
	For evaluation, we use beam search with a beam size of $k=4$ and length penalty $\alpha$=0.6.
	Additionally, we use 6 encoder and decoder layers. In each layer, we employ  $d_{model}=1024$ hidden size, 16 attention-heads, 4096 feed forward inner-layer dimensions, and $P_{dropout}$=0.1.
	Our settings are close to \emph{transformer\_big} setting as defined in \cite{vaswani2017attention}.
	We employ three Titan Xp GPUs to train English-to-German translation and one GPU for Chinese-to-English translation pairs. In addition, we average the last 20 checkpoints to get the final model for English-to-German but do not perform checkpoint averaging for Chinese-to-English.
	
	We evaluate the final translation quality with case-insensitive BLEU \cite{papineni2002bleu} for Chinese-to-English and with case-sensitive BLEU for English-to-German. Significance test is performed using the pairwise re-sampling approach \cite{koehn2004sig}.
	
	\subsubsection{Translation Systems}
	
	We use {\bf BI-RNMT} to denote our proposed synchronous bidirectional inference model implemented in LSTM-based recurrent neural machine translation. The proposed synchronous {\bf b}idirectional {\bf i}nference model {\bf f}or {\bf T}ransformer is named {\bf BIFT}{\footnote{Our code is freely available in github https://github.com/ZNLP/sb-nmt}}.
	
	We compare the proposed models against the following state-of-the-art NMT systems:
	
	\begin{itemize}
		\item {\bf RNMT}~\cite{wu2016google}: it is a  state-of-the-art LSTM-based NMT system with the same setting as {\bf BI-RNMT}.
		\item {\bf RNMT (R2L)}: it is a variant of RNMT and produces translations from right to left.
		\item {\bf Transformer}: it is the state-of-the-art machine translation system with self-attention mechanism using the default left-to-right generation~\cite{vaswani2017attention}.
		\item {\bf Transformer (R2L)}: it is a variant of Transformer which performs translation in a right-to-left manner.
		\item {\bf Rerank-NMT}: following \cite{liu2016agreementa}, we first run beam search for L2R and R2L inference models independently to obtain two k-best lists, and then re-score the union of these two k-best lists. This method assumes that some source sentences are appropriate to translate from left to right, while others are better to translate from right to left.
		\item {\bf ABD-NMT}: it is an asynchronous bidirectional inference model for NMT that performs L2R inference with the results generated by R2L inference model~\cite{zhang2018asynchronous}. During inference, two-pass decoding scheme is employed. First, the R2L inference model generates the backward hidden states and corresponding translation results. Then, ABD-NMT optimizes the L2R inference with the helpf of the backward hidden states.
	\end{itemize}
	
	For fair comparison, Rerank-NMT and ABD-NMT are all reimplemented based on strong Transformer models.
	
	\subsection{Abstractive Summarization}
	
	\subsubsection{Dataset}
	
	Abstractive sentence summarization is a task that generates a title-like summary for a long sentence. Our {\bf training data} is a (text, summary) parallel corpus from the Annotated English Gigaword dataset \cite{napoles2012annotated,rush2015neural}. It contains about 3.8M text-summary pairs for training and 189K pairs for validation. The encoder and decoder share the same vocabulary of about 90K word types.
	
	For the {\bf test set}, we use both DUC 2004 and the English Gigaword.  In the test set of DUC 2004, there are 500 examples and each example pairs a document with four different human-written reference summaries.  For the test set of the English Gigaword, we employ the same randomly selected subset of 2000 text-summary pairs as \cite{rush2015neural,zhou2017selective}.
	
	\subsubsection{Training and Evaluation Details}
	
	For both LSTM-based architecture and Transformer framework, we use the same model settings as neural machine translation.
	
	For evaluation, we use ROUGE \cite{lin2004rouge} as the metric. ROUGE measures the quality of summary
	by computing overlapping lexical units, such as unigram, bigram, trigram, and longest common subsequence (LCS). It becomes the standard evaluation metric for DUC shared tasks and popular
	for summarization evaluation. Following previous work, we use ROUGE-1 (unigram), ROUGE-2 (bigram) and ROUGE-L (LCS) as our evaluation metrics in the reported experimental results.
	
	\subsubsection{Summarization Systems}
	
	We compare our proposed model with the following state-of-the-art baselines.
	
	{\bf ABS}: \cite{rush2015neural} first proposed the abstractive summarization task and used an attentive CNN encoder and NNLM (neural network language model) decoder to perform this task.
	
	{\bf LSTM-Sum}: it is the abstractive summarization system with the same architecture as RNMT in which the encoder and decoder are both LSTM-based recurrent neural networks.
	
	{\bf Feats2S}: it is also a RNN encoder-decoder model using gated recurrent unit (GRU) \cite{bahdanau2015neural} and provide more features (e.g. POS and NER) to enrich the encoder \cite{nallapati2016abstractive}.
	
	{\bf Selective-Enc}: \cite{zhou2017selective} proposed a selective mechanism to selecting important information from encoder before generating summary.
	
	{\bf Transformer}: it is a Transformer model which is applied to the abstractive sentence summarization task.
	
	\section{Results and Analysis}
	
	\subsection{Machine Translation}
	
	\subsubsection{Overall Translation Quality}
	
	\begin{table*}
		\centering
		\begin{tabular}{l|cccc|cc}
			\hline
			Model                     &MT03   &      MT04   &      M05     &      MT06       & AVE      &  $\Delta$ \\
			\hline
			\hline
			RNMT                & 42.07     &    43.40     &       40.73    &    41.11        &     41.83   & -   \\
			RNMT (R2L)                &  41.47    &    43.13      &     40.62    &   40.94       &     41.54    &   -0.29\\
			{\bf BI-RNMT}               &  {\bf 43.50}    &    {\bf 43.98}      &     {\bf 41.37}    &   {\bf 42.48}       &     {\bf 42.83}    &   {\bf +1.00}\\
			\hline
			\hline
			Transformer        &  47.63    &    48.32     &       47.51     &   45.31       &     47.19    & -\\
			Transformer (R2L)     &  46.79     &    47.01      &      46.50   &   44.13       &     46.11    &  -1.08\\
			Rerank-NMT             &  48.23       & 48.91    &     48.73    & 46.51       &     48.10 &  +0.91 \\
			ABD-NMT            &  49.47     &       48.01        &       48.19        &      47.09    &   48.19    & +1.00 \\
			{\bf BIFT}             & {\bf 51.87}        &     {\bf 51.50}   &        {\bf 51.23}    &    {\bf 49.83}    &  {\bf 51.11}  & {\bf +3.92}   \\
			
			\hline
		\end{tabular}
		\caption{Translation quality for Chinese-to-English tasks using case-insensitive BLEU scores.
			Our model {\bf BI-RNMT} and {\bf BIFT} are respectively significantly better than corresponding baselines  (p $<$ 0.01).
		} \label{table-zh-en}
	\end{table*}
	
	Table~\ref{table-zh-en} reports the translation performance of different systems on the Chinese-English task. The results are mainly divided into two parts. The first part in this table shows the BLEU scores of the systems based on LSTM framework while the second part gives the results of various systems based on the Transformer architecture.
	
	Comparing the baselines using different architectures, we can easily see that the self-attention based Transformer remarkably outperforms the LSTM-based RNMT, with the average improvement of 5.36 BLEU points (47.19 vs. 41.83), suggesting the superiority of the Transformer.
	
	As conventional decoding performs left to right, a question may arise that which inference direction is better. It is easy to find from table~\ref{table-zh-en} that the right-to-left decoding performs worse than the left-to-right style no matter which neural network architecture is adopted (RNMT (R2L) vs. RNMT, Transformer (R2L) vs. Transformer). Specifically, the gap under the Transformer architecture is much bigger (1.08 vs. 0.29), indicating that Transformer is more sensitive to the inference direction.
	
	We also investigate previous methods that take advantage of two inference directions. The second part in table~\ref{table-zh-en} shows that both the reranking approach Rerank-NMT \cite{liu2016agreementa} and the asynchronous bidirectional decoding method ABD-NMT \cite{zhang2018asynchronous} can get a significant improvement over the strong Transformer baseline. The average gains can be up to 1.0 BLEU point (48.19 vs. 47.19), indicating that L2R decoding and R2L decoding can be complementary to each other.
	
	We go step further to exploit synchronous bidirectional inference that makes full use of L2R and R2L decoding. It is obvious to see from table~\ref{table-zh-en} that our proposed method performs best. The first part in table~\ref{table-zh-en} says that the synchronous bidirectional inference model under LSTM framework {\bf BI-RNMT} can obtain an average improvement of 1.0 BLEU point over RNMT. The second part demonstrates that our synchronous bidirectional inference model under the Transformer architecture {\bf BIFT} achieves promising BLEU gains and the gap can be as large as 3.92 BLEU points on average. The remarkable improvements suggest that compared to asynchronous bidirectional decoding, our synchronous bidirectional inference can better explore the history and future contexts on the target side.
	
	\begin{table}
		\centering
		\begin{tabular}{l|c}
			\hline
			Model              &    TEST  (WMT 14)              \\
			\hline
			\hline
			GNMT~\cite{wu2016google}                 &         24.61            \\
			Conv~\cite{gehring2017convolutional}     &           25.16          \\
			AttIsAll~\cite{vaswani2017attention}     &      28.40        \\
			\hline
			\hline
			RNMT	& 22.85 \\
			RNMT (R2L)	& 22.17 \\
			{\bf BI-RNMT}	& 23.97 \\
			\hline
			\hline
			Transformer\footnotemark[8]   &       27.72     \\
			Transformer (R2L)  &      27.13          \\
			Rerank-NMT          &     27.81      \\
			ABD-NMT           &     28.22      \\
			\hline
			\hline
			{\bf BIFT}       &     {\bf 29.21}     \\
			\hline
		\end{tabular}
		\caption{Translation results on WMT14 English-to-German task using case-sensitive BLEU.
		} \label{table-en-de}
	\end{table}
	\footnotetext[8]{The BLEU score of the Transformer model is reproduced in our hardware environment and is slightly lower than AttIsAll \cite{vaswani2017attention}. \cite{chen2018best} also reported that their reproduction is lower than their original result in \cite{vaswani2017attention}. In our experiments, we use only 3 GPUs for English-to-German, whereas \cite{vaswani2017attention,chen2018best} adopted TPUs for model training.}
	
	Similar phenomena can be observed from the English-German translation results as shown in Table~\ref{table-en-de}. The finding is that {\bf BI-RNMT} outperforms RNMT with 1.12 BLEU points. {\bf BIFT} performs best among all the systems including GNMT~\cite{wu2016google}, Conv~\cite{gehring2017convolutional} and AttIsAll~\cite{vaswani2017attention}. In addition, {\bf BIFT} achieves the state-of-the-art performance of 29.21 on the same dataset. Considering that only one reference is available for English-to-German translation, the improvements are very promising.
	
	\subsubsection{Model Size and Efficiency}
	
	Our synchronous bidirectional inference model is slightly complicated than conventional L2R or R2L inference model. It is interesting to figure out the model size and efficiency of our system compared to other baseline systems. Table~\ref{model-size} reports the corresponding statistics of different NMT models. The model size denotes the total number of network parameters. Since the synchronous bidirectional inference model only introduces one parameter $\lambda$ in Equation~\ref{history-future-comb} for BIFT, the model size is the same as that of the Transformer. In contrast, Rerank-NMT has double the number of parameters compared to the Transformer because it requires two individual encoder-decoder models for L2R and R2L decoding respectively. As for the asynchronous bidirectional decoding model, ABD-NMT shares one encoder and has two decoders, and thus contains more than a half parameters against the Transformer baseline.
	
	\begin{table}
		\centering
		\begin{tabular}{l|c|cc}
			\hline
			\multirow{2}{*}{Model} &  \multirow{2}{*}{Model Size}   &   \multicolumn{2}{c}{Efficiency} \\
			&                               &    \emph{Train}   & \emph{Test}     \\
			\hline
			\hline
			Transformer       &  207.8M &    2.07       &  19.97 \\
			Transformer (R2L)  &  207.8M &    2.07          &  19.81  \\
			Rerank-NMT         &  415.6M&    1.03       & 6.51\\
			ABD-NMT            &  333.8M &    1.18          & 7.20 \\
			\hline
			\hline
			{\bf BIFT}          &  207.8M &      1.26   &    17.87 \\
			\hline
		\end{tabular}
		\caption{Comparison results of model size, training and testing efficiency. \emph{Train} denotes the number of global training steps processed per second at the same batch-sized sentences; \emph{Test} indicates the amount of translated sentences in one second.} \label{model-size}
	\end{table}
	
	The {\em Train} column shows the number of global training steps per second. Because the training procedure of BIFT needs to match both of the L2R and R2L references, it takes more time to converge. However, regarding the decoding efficiency, our synchronous bidirectional inference model performs on par with the Transformer baseline and is only 10\% slowdown (17.87 vs. 19.97 sentences per second), whereas Rerank-NMT and ABD-NMT are much slower. The statistics suggest that BIFT is acceptable regarding the decoding efficiency.
	
	\subsubsection{Performance Trends on Sentence Length}
	
	\begin{figure}[!t]
		\centering
		\includegraphics[scale=1.0]{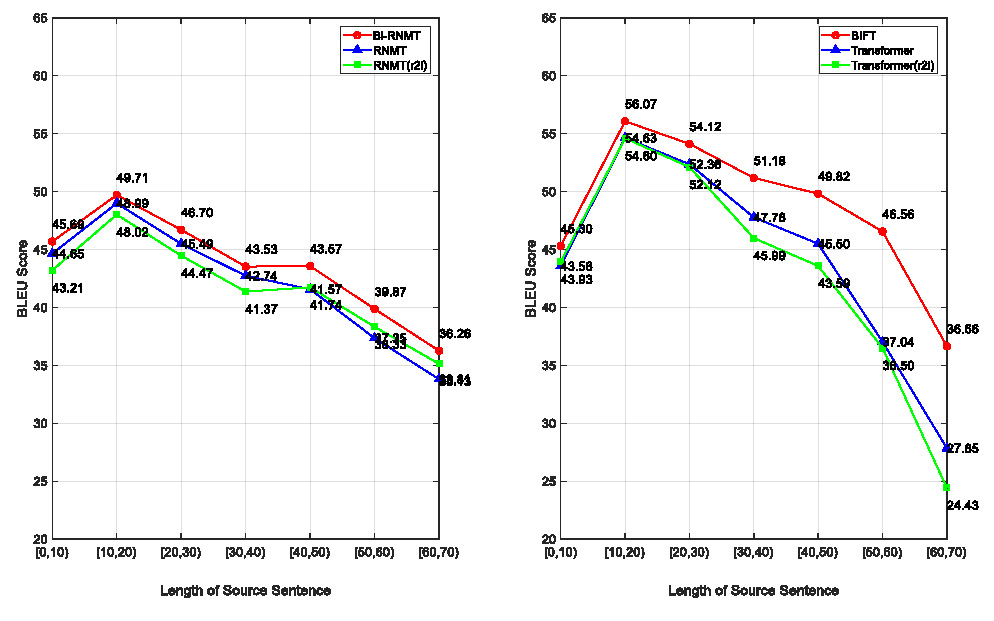}
		\caption{BLEU score trends for different intervals of sentence length.}
		\label{bleu-length}       
	\end{figure}
	
	In previous sections, we argued that BI-RNMT and BIFT can utilize both of the history and future contexts during translation. A natural question may arise that whether our proposed model would perform much better on the long sentences. To answer this question, we group the source sentences of similar lengths in the test set and calculate the corresponding BLEU scores for each length interval.
	
	Fig.~\ref{bleu-length} displays the statistics. The left part of Fig.~\ref{bleu-length} represents the results of the systems using LSTM-based framework. The right one shows the statistics of Transformer-based systems. Overall, no matter which architecture is adopted, our proposed models (BI-RNMT and BIFT) are superior to baselines over sentences with all different lengths. Generally, the gap becomes bigger and bigger when the length grows. Comparing the two architectures, we find that the Transformer-based BIFT excels in long sentence translation. The results indicate that our proposed synchronous bidirectional inference model are better at translating long sentences with the help of both the history and future contexts during decoding.

	\subsubsection{Translation Precision over Different Positions}
	
	We mentioned in introduction that L2R inference is good at predicting prefix while R2L inference is adept at suffix prediction. We may wonder that how does our synchronous bidirectional inference model perform on prefix and suffix prediction. Furthermore, which part of the translation sentence will be improved most, the prefix, middle part or the suffix?
	
	\begin{table*}
		\centering
		\begin{tabular}{l|c|c}
			\hline
			\bf Model & \bf First Four & \bf Last Four \\ \hline
			RNMT & 36.35\% & 31.64\% \\
			RNMT (R2L) &   31.22\% & 34.01\% \\
			{\bf BI-RNMT} &   \bf 36.88\% & \bf 34.65\% \\\hline \hline
			Transformer  & 40.21\% & 35.10\% \\
			Transformer (R2L) &   35.67\% & 39.47\% \\
			Rerank-NMT &   38.98\% & 38.91\% \\
			ABD-NMT &   38.36\% & 38.11\% \\
			{\bf BIFT}  &   \bf 40.89\% & \bf 40.08\% \\
			\hline
		\end{tabular}
		\caption{Matching accuracy of the first and last four tokens between model predictions and references in NIST Chinese-English machine translation tasks for different NMT systems.
		} \label{match-acc-new}
	\end{table*}
	
	\begin{figure}[!t]
		\centering
		\includegraphics[scale=1.0]{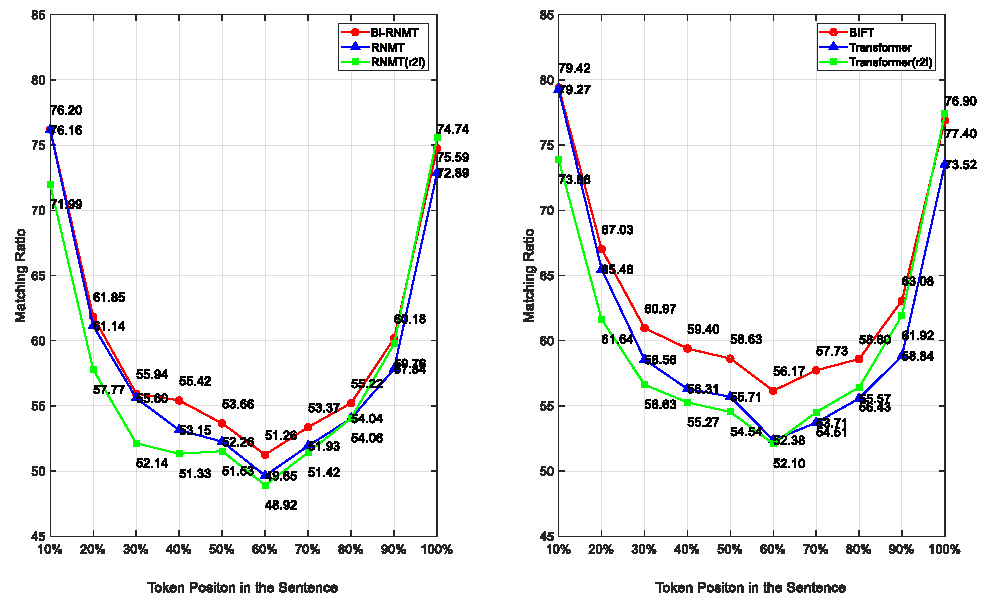}
		\caption{Match precision of translation tokens for different positions.}
		\label{match-position}       
	\end{figure}
	
	To figure out these questions, we first analyze the prediction precision of the first-four and last-four tokens of the translation compared to the references for different NMT systems. Table~\ref{match-acc-new} reports the comparison results. Obviously, the proposed synchronous bidirectional inference model performs best for matching precision of both the first-four and last-four tokens, showing the superiority of our methods.
	
	Then, we go step further and conduct a deep analysis. We divide each translation hypothesis and its reference into 10 equal parts and calculate the average word prediction accuracy for each part. In this way, we attempt to investigate the contribution of the synchronous bidirectional inference model over different positions. Fig.~\ref{match-position} illustrates the comparison results. It is interesting to see from this figure that both ends of the translation hypothesis are much easier to predict. In contrast, the prediction accuracy of the middle part is much lower, suggesting more demands of both the history and future contexts. Our proposed BI-RNMT and BIFT facilitate the usage of the left and right predictions, leading to large improvements over the middle part (40-80\% in Fig.~\ref{match-position}).

	\subsubsection{Two-pass Training vs. Fine-tuning}
	In this subsection, we attempt to investigate the effects of different parameter optimization strategies for our synchronous bidirectional inference model. We employ BIFT and Chinese-to-English translation task to compare between the two-pass training strategy and the fine-tuning strategy. In the fine-tuning step, we randomly choose 10\% source sentences of the training data. Table~\ref{trainstrategy} reports the comparison results.
	
	As shown in the table, we observe that both training strategies can remarkably improve the translation performance compared to the Transformer baseline. Although the fine-tuning strategy is not as powerful as the two-pass training strategy, it can still achieve a big improvement of 2 BLEU points in average over the strong baseline. Considering that the fine-tuning strategy is much easier and cheaper for system deployment, we believe this strategy will be more popular than the two-pass strategy.
	
	\begin{table*}
		\centering
		\begin{tabular}{l|cccc|cc}
			\hline
			Model                     &MT03   &      MT04   &      M05     &      MT06       & AVE      &  $\Delta$ \\
			Transformer        &  47.63    &    48.32     &       47.51     &   45.31       &     47.19    & -\\
			{\bf BIFT (two-pass)}             & {\bf 51.87}        &     {\bf 51.50}   &        {\bf 51.23}    &    {\bf 49.83}    &  {\bf 51.11}  & {\bf +3.92}   \\
			{\bf BIFT (fine-tuning)}             & {\bf 50.76}        &     {\bf 49.72}   &        {\bf 48.32}    &    {\bf 47.91}    &  {\bf 49.18}  & {\bf +1.99}   \\
			\hline
		\end{tabular}
		\caption{Translation results of different training strategies for our synchronous bidirectional inference model on Chinese-to-English tasks.
		} \label{trainstrategy}
	\end{table*}
	
	\subsection{Abstractive Summarization}
	
	\subsubsection{Summarization Quality}
	\begin{table}
		\centering
		\begin{tabular}{l|c|c|c|c|c|c}
			\hline
			\multirow{2}{*}{Model}    &   \multicolumn{3}{c|}{DUC-2004}  &   \multicolumn{3}{c}{English Gigaword} \\ \cline{2-7}
			&   R1  & R2 & R-L     &   R1 & R2   & R-L   \\
			\hline
			ABS   &     26.55 & 7.06 & 22.05 & 29.55 & 11.32 & 26.42 \\
			Feats2s   &     28.35 & 9.46 & 24.59 & 32.67 & 15.59 & 30.64 \\
			Selective-Env &   \bf 29.21 & 9.56 & 25.51 & \bf 36.15 & \bf 17.54 & \bf 33.63 \\
			\hline
			\hline
			RNMT  &     28.22 & 10.21 & 25.14 & 34.54 & 16.85 & 32.32 \\
			{\bf BI-RNMT} &   29.05 & \bf 10.90 & \bf 26.05 & 35.47 & 17.62 & 32.90 \\
			\hline
			\hline
			Transformer   &     28.09 & 9.52 & 24.91 & 34.12 & 16.04 & 31.46 \\
			{\bf BIFT} &   29.17 & 10.30 & \bf 26.05 & 35.68 & 17.39 & 32.89 \\
			\hline
		\end{tabular}
		\caption{Abstractive summarization quality on DUC 2004 and English Gigaword for different methods.} \label{table-sum}
	\end{table}
	
	Abstractive sentence summarization is another well-known testbed for sequence to sequence learning. We then apply our synchronous bidirectional inference model into this task. Table~\ref{table-sum} presents the results of different systems over two test sets DUC-2004 and English Gigaword.
	
	The first three rows show the performances of previous state-of-the-art abstractive summarization models. Among these three baselines, both Feats2s and Selective-Env aim at improving the summarization quality by enhancing the representation learning of the encoder. Selective-Env augments the encoder with key information selection performs best.
	
	In contrast, our method attempts to improve the decoder (inference module) by enabling synchronous bidirectional decoding. The last four rows in Table~\ref{table-sum} demonstrate that the LSTM-based RNMT stably outperforms the self-attention based Transformer, which is quite different from that for neural machine translation in which Transformer is the better one. After applying our synchronous bidirectional inference model, BI-RNMT and BIFT respectively achieve significantly better results on two test sets over RNMT and Transformer, despite that these two models cannot outperform Selective-Env on the Gigaword test set. The reason behind may be that we just apply BI-RNMT and BIFT into abstractive summarization without any special adaptation processing. From another perspective, the two kinds of the models handle encoder and decoder respectively, and can be complementary to each other. Nevertheless, the statistics given in the table further show the effectiveness of our synchronous bidirectional inference model beyond machine translation.
	
	\subsubsection{Some Examples}
	
	\begin{table*}[!ht]
		\centering
		\begin{tabular}{|p{1.5cm}|p{9.5cm}|}
			\hline
			Input &  resident nelson mandela acknowledged saturday the african national congress violated human rights during apartheid , setting him at odds with his deputy president over a report that has divided much of south africa .\\
			\hline
			Reference & mandela acknowledges human rights violations by african national congress \\
			\hline
			Transformer & mandela acknowledges human rights violation at odds with deputy president   \\
			\hline
			Transformer (R2L) & mandela says south africa violated human rights \\
			\hline
			BIFT &  mandela says south african national congress violated human rights  \\
			\hline
			\hline
			Input &  the new york times said in an editorial on monday , oct. UNK : since the deadly bombing of two american embassies in africa in august , there has been a troubling accumulation of evidence that the state department inexplicably ignored warnings of possible terrorist attacks against the installations .\\
			\hline
			Reference & editorial claims state department ignored warnings of terrorist \\
			\hline
			Transformer & new york times says evidence of terror attacks is ignored   \\
			\hline
			Transformer (R2L) & new york times warns of possible terrorist attacks on u.s. embassies in africa \\
			\hline
			BIFT & new york times says state department ignored terror warnings  \\
			\hline
		\end{tabular}
		\caption{Abstractive summarization examples comparing our synchronous bidirectional inference model to other baselines.} \label{example}
	\end{table*}
	
	To better understand the models, we further investigate some specific examples which are listed in Table~\ref{example}. For each example, the input is a long sentence and the output is a title-like summary. 
	
	In the first example, {\em human rights} and {\em african national congress} are two key contents. The baseline Transformer fails to generate {\em african national congress} in the tail part and Transformer (R2L) neglects {\em natrail congress} in the head part. In contrast, our model BIFT renders all of the key points.
	
	As for the second example, the baselines have made similar mistakes. Transformer omits {\em by state department} at the end, and the summary generated by Transformer (R2L) expresses the wrong meaning due to the absence of {\em says state department fails to} after {\em new york times}. However, our model BIFT can generate the summary with correct and complete contents. The examples demonstrate the superiority of our synchronous bidirectional inference model over the modeling of both history and future contexts.
	
	\section{Related Work}
	
	This work addresses synchronous bidirectional inference for sequence to sequence learning tasks, aiming to take full advantage of history and future predictions on the output. Generally, the related work can be divided into two categories, namely bidirectional inference and future context usage.
	
	Bidirectional inference is well studied in sequential labeling tasks \cite{toutanova2003feature,tsuruoka2005bidirectional,shen2007guided}, in which each input token corresponds to an output label and the output shares the same length with the input. In general, there are $2^{n-1}$ decomposition ways of the conditional probability $p(y_0^{n-1}|x_0^{n-1})$ for an $n$-token input sequence, since each token is predicted after the left one (from left to right) or the right one (from right to left). In this way, bidirectional inference is not difficult for sequential labeling. However, it is not trivial to leverage bidirectional inference for sequence generation problems mainly due to the length nondeterminacy of the output. \cite{liu2016agreementa,liu2016agreementb,zhang2018asynchronous} added agreement constraints to enforce L2R inference output to be consistent with R2L inference output for sequence generation tasks. \cite{serdyuk2018twin} proposed the twin network that encourages the target hidden states of the L2R and R2L inferences at the same position to be as close as possible to predict the same token during training. Recently, \cite{zhang2018asynchronous} introduced an asynchronous bidirectional inference model for neural machine translation. They first obtained the translation hypothesis using R2L inference and then optimized the L2R inference model with the help of the R2L inference result. Despite of performance improvement, all these studies require two individual inference models, making the architecture more complicated. Furthermore, the interactions between L2R and R2L inferences are not adequate. Taking the asynchronous bidirectional inference model for example,  L2R model can utilize the information of R2L model but R2L inference cannot use the L2R predictions. In contrast, our synchronous bidirectional inference model has only a single decoder in which L2R and R2L inferences interact with each at each decoding time step.
	
	Using future contexts has drawn more and more attention in sequence prediction tasks. Intuitively, R2L inference model can be employed to re-rank the $n$-best hypotheses of the L2R inference model, preferring balanced output \cite{sennrich2016edinburgh,sennrich2017university,hoang2017towards,tan2017xmu,deng2018alibaba,liu2018comparable}. To use the future context which is unavailable in conventional inference model, \cite{bahdanau2017actor,he2017decoding,li2017learning} proposed the reinforcement learning methods to estimate the possible future information. To mimic the human cognitive behaviors, \cite{xia2017deliberation} presented a deliberation network, which leverages the global information with the help of both forward and backward predictions in sequence generation through a deliberation process. \cite{zheng2018modeling} introduced two additional recurrent layers to model translated past contents and untranslated future contents.
	They much improved the sequence generation quality with the cost of model complexity. They either employed two-pass decoding strategy or added more layers to the original network. Compared to this kind of work, our proposed model uses a smart way to exploit both history and future predictions by allowing L2R and R2L inferences to perform in parallel but interactively. In our previous work \cite{zhou2019synchronous}, we address the bidirectional decoding for neural machine translation. In this current work, we generalize the decoding model into synchronous bidirectional inference for the general sequence-to-sequence models (LSTM and Transformer) and general sequence generation tasks (translation and summarization). We further propose and investigate two optimization strategies to learn network parameters.

	\section{Conclusion and Future Work}
	This work proposes a synchronous bidirectional inference model for sequential generation tasks. We first presented a synchronous bidirectional beam search algorithm for sequence generation, in which left-to-right and right-to-left decoding perform in parallel but interactively. We have exploited the usage of synchronous bidirectional inference model on both LSTM-based and Transformer-based seq2seq architectures. We have also proposed and investigated two parameter optimization strategies. The comprehensive experiments on machine translation and abstractive summarization have demonstrated that our proposed synchronous bidirectional inference model remarkably outperforms the strong baselines. The deep analysis further shows that our model can indeed take full advantage of both history and future predictions during inference.
	
	
	In the future work, we plan to apply our synchronous bidirectional inference model to other sequential generation tasks, such as question answering, chatbot and image caption.

\bibliography{elsarticle-template}

\end{document}